# Surveillance of COVID-19 Pandemic using Hidden Markov Model

Shreekanth M. Prabhu[1] & Natarajan Subramaniam[2]

[1]Professor, Department of Information Science and Engineering, CMR Institute of Technology, Bengaluru, India.

Address: 132 AECS Layout ITPL Main Road, Kundalahalli Bengalurur- 560037, India

email: shreekanthpm@gmail.com, shreekanth.p@cmrit.ac.in  ph: +919880479148

[2]Professor, Department of Computer Science and Engineering, PES University, Bengaluru, India

Address: 100 Feet Outer Ring Road, BSK III Stage, Banglaore - 560085

Email: natarajan@pes.edu, ph: +91994528022

Corresponding Author Email: shreekanthpm@gmail.com

# Surveillance of COVID-19 Pandemic using Hidden Markov Model

## Abstract

COVID-19 pandemic has brought the whole world to a stand-still over the last few months. In particular the pace at which pandemic has spread has taken everybody off-guard. The Governments across the world have responded by imposing lock-downs, stopping/restricting travel and mandating social distancing. On the positive side there is wide availability of information on active cases, recoveries and deaths collected daily across regions. However, what has been particularly challenging is to track the spread of the disease by asymptomatic carriers termed as super-spreaders. In this paper we look at applying Hidden Markov Model to get a better assessment of extent of spread. The outcome of such analysis can be useful to Governments to design the required interventions/responses in a calibrated manner. The data we have chosen to analyze pertains to Indian scenario

## 1. Introduction

COVID-19 is the name given by WHO to the disease caused by SARS Cov-2 virus. Whereas the death rates of earlier SARS virus were higher than Cov-2 virus, as a pandemic COVID-19 has spread far more speedily infecting Lakhs of people.

Over the last few decades and centuries there has been incidence of Pandemics and losses of human lives on large scale from time to time. However, the terminology around Pandemic can be confusing. An endemic is something that happens to a particular people or region. For example, Malaria is endemic to certain countries. An outbreak is a greater-than-anticipated increase in the number of endemic cases. It can also be a single case in a new area. Thus, an outbreak can happen in regions where a given disease is endemic or in totally new regions. If it's not quickly controlled, an outbreak can become an epidemic. An epidemic is often localized to a region, but the number of those infected in that region is significantly higher than normal. When an epidemic spread world-wide due to travel or other reasons it is termed pandemic. It is the geographic spread that is vital to spread of pandemic [1].

Naturally the first response to arrest COVID-19 spread should have been ban of travel from and to regions where the disease was endemic. However due to complex factors which played out that did not happen in a timely manner and the pandemic has been spreading since early January when it was publicly acknowledged world-wide. On top of that people travelling in from only certain countries were screened while infection had already spread to lot more countries. The travellers who did not display symptoms were asked to be in isolation and those who displayed symptoms were quarantined. This again did not happen to the degree that was needed.

Once an outbreak happens in a new region, expected response was to arrest further spread to more people. This however could only be done retroactively by tracing contacts a newly detected person may have encountered and testing them. The Governments had limited capacity to test and they needed to manage them judiciously. This resulted in continued spread of the disease in a latent manner. The COVID-19 spread from person to person via droplets that got dispersed when a person was coughing, sneezing or even talking. It was also suspected that the virus spread through surfaces and on certain surfaces the virus lasted for couple of days or more. Many of these conclusions were based on simulation studies and not necessarily attested from field studies.

In addition to pervasive mode of spreading another challenge with COVID-19 is that many infected people do not display any symptoms for long periods. Table 1 below compares COVID-19 with other viruses [2]. You can see that COVID-19 has the longest incubation period during which virus can be contagious i.e. even before the symptoms start. A person will continue to be contagious with onset of symptoms and may be for some more time even after recovery.

It was also seen that those with pre-existing health problems were more vulnerable. On an average the disease impacted the elderly lot more. In the case of any new virus, medical communities take a few months in the least to come up with a vaccine that can build immunity among people. Further, vaccines are of no use to people already sick. To treat them suitable antiviral drugs are needed which can shorten the time people are unwell. They may need to be taken early on and may not work in every case. There has been vocal community of experts urging Governments to test, test and test. But considering the size of populations involved and the scale of effort required it has proved impractical. At the same time, it is worthwhile to test only in regions and among communities where there is a possible outbreak. Inviting people to common testing facilities can itself expose them to chance of infection. Even though there has been debates about herd immunity by allowing the virus to spread among healthier people, making such calls has been rather highly risky. Finally, it is not health experts but the politicians and Governments who have to take responsibility for their actions. General information on COVID-19 is presented in a lucid manner by Robert Roy Britt [3].

Table 1: Comparing COVID 19 with other viruses

|  | **COLD** | **FLU** | **NOROVIRUS** | **COVID-19** |
|---|---|---|---|---|
| **Incubation Period** | 1-3 days | 1-4 days | A few hours | 2-14 days |
| **Symptom Onset** | Gradual | Abrupt | Abrupt | Gradual |
| **Typical illness duration** | 7-10 days | 3-7 days | 1-2 days | undetermined |
| **Symptoms** | | | | |
| **Sore Throat** | **Common** | Some Times | *Rare* | Some Times |
| **Sneezing** | **Common** | Some Times | *Rare* | *Rare* |
| **Stuffy, runny nose** | **Common** | Some Times | *Rare* | Some Times |
| **Cough, chest discomfort** | Some Times | **Common** | *Rare* | **Common** |

| Fatigue, weakness | Some Times | **Common** | Some Times | Some Times |
| --- | --- | --- | --- | --- |
| **Fever** | *Rare* | **Common** | Some Times | **Common** |
| **Aches** | *Rare* | **Common** | Some Times | Some Times |
| **Chills** | *Rare* | **Common** | Some Times | Some Times |
| **Headache** | *Rare* | **Common** | Some Times | *Rare* |
| **Shortness of breath** | *Rare* | *Rare* | *Rare* | **Common** |
| **Nausea** | *Rare* | *Rare* | **Common** | *Rare* |
| **Vomiting** | *Rare* | *Rare* | **Common** | *Rare* |
| **Diarrhea** | *Rare* | *Rare* | **Common** | *Rare* |
| **Stomach Pain** | Rare | Rare | **Common** | *Rare* |

Thus, when Science is found to be wanting, we should turn to Mathematics for solutions. In this paper we propose application of Hidden Markov Model to assess the spread of infection in communities by making use of publicly available data on active cases, recoveries and deaths. We make use of data published by Ministry of Health and Family Welfare, Government of India. Currently India as of early May 2020, is possibly able to flatten the curve but yet to see a dip in infections [4]. The situation can change unpredictably as more people are allowed to travel back to India or across India.

We hope that this work can help Governments calibrate interventions/responses using a relatively simple model. We also outline strategies for building more complex models to deliver fine-grained analysis.

The remaining part of the paper is structured as follows. Section 2, covers Literature Survey where we cover literature on surveillance of past pandemics followed by literature related to COVID-19. Section 3 covers the Proposed Methodology for COVID-19 Pandemic Surveillance using Hidden Markov Model. Section 4 presents Results on application of the proposed methodology to Indian context. In Section 5, we make a list of Recommendations for better surveillance of COVID-19 Pandemic in India. Section 6 concludes the paper.

## 2. Literature Survey

The three pandemics in the twentieth century killed between 50 and 100 million people and the 2009 H1N1 "Swine Flu" exposed vulnerabilities that we still have to influenza epidemics. David Hutton [5] has done a review of Operations Research Tools and Techniques Used for Influenza Pandemic Planning. He covers instances where Simulation modelling is used in as a way of representing the real world and being able to estimate the impact of interventions and to improve their performance.

Among the simulation models, system dynamics approach can capture how the changing level of infection and immunity in the population affects the spread of future infections in the population. System dynamics modelling techniques have been applied to pandemic influenza preparedness problems as varied as social distancing, vaccination, antiviral treatment, and portfolio analysis of interventions. Larsen [6] showed with his simple model that the people with high rates of contact drove the initial growth of the epidemic and that targeting social distancing (reducing contact rates) to the correct subpopulations with high contact rates can prevent the epidemic with limited disruption to the remaining population. Nigmatulina and Larson [7] built off that simple dynamic compartmental model of influenza to include multiple interconnected communities. Each community has citizens that interact with each other and is connected to other communities by a few travellers each day. This might represent towns near each other. They used this model to compare the impact of vaccination and travel restrictions. They also find that travel restrictions will not be effective: only a complete 100 % travel restriction would stop or significantly slow transmission between communities. The above analysis [7] and the one by Larson alone [6] use very simple models that can be implemented in spreadsheets, but they provide very powerful policy insights. System Dynamics models also have been used to evaluate antiviral use [8] as well as vaccine delivery mechanisms [9] and optimal allocation of health-care resources. Discrete event simulations are used to manage distribution systems. Agent based models have been used where more elaborate data is available to decide on policy choices such as who should be vaccinated first [10]. Vaccine selection problem can be formulated as optimization problem when one has to choose between releasing an available vaccine or waiting for better vaccine. Along the same lines, Hutton refers to cases where decision analysis, game theory and supply chain analysis are used to manage health-care resources.

Rath et al. [11] made use of Hidden Markov Model for automated detection of Influenza epidemic. He builds on Serfling's method that is commonly used for Influenza Surveillance. The method uses cyclic regression (to address cyclical data owing to seasonality of epidemics) to model the weekly proportion of deaths from pneumonia and influenza and to define an epidemic threshold that is adjusted for seasonal effects. That approach however suffers from several shortcomings, such as the need for non-epidemic data to model the baseline distribution, and the fact that observations are treated as independent and identically distributed. This can be handled by modeling the data as time series and use Hidden Markov Model to segment it as non-epidemic and epidemic phases. Rath et al. in their paper have used exponential distribution for non-epidemic phase and gaussian distribution for epidemic phase. This is a refinement over the work done by Le Strat & Carrat [12] who made use of only Gaussian Distributions.

## 2.1 Literature related to COVID-19 Pandemic

COVID-19 Pandemic has been an active area of research last few months. There have also been large number of articles in main-stream, social media and as part of professional health-care literature. Here we cover literature related to testing procedures and pharmaceutical interventions, non-pharmaceutical interventions followed by mathematical analysis of forecasts globally and in India. Finally, we look at aspects related to surveillance and social implications.

Nandini Sethuraman et al. [13] emphasize on importance of timing to get the correct results in Real Time Reverse Transcription-Polymerase Chain Reaction (RT-PCR) Tests that are used to detect SARS-COV-2 RNA as illustrated in figure below. Thus, there is a very real possibility that a sizable number of COVID-19 cases are not detected on testing.

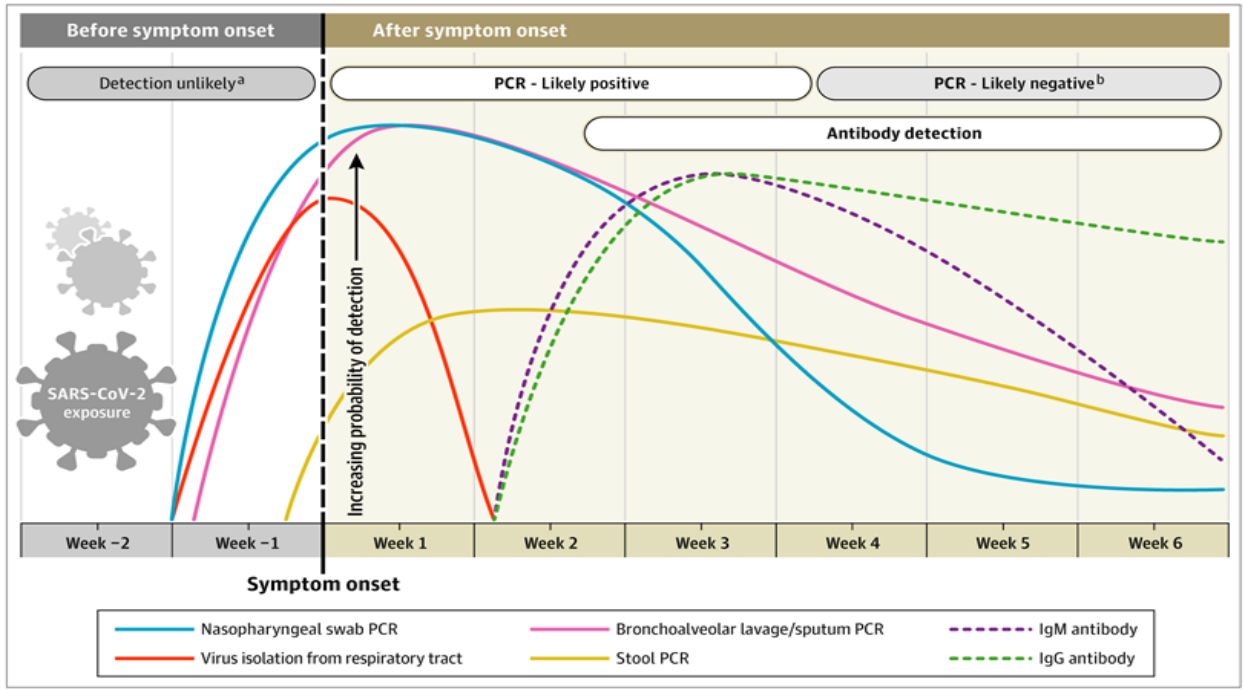

**Figure 1: Estimated Variation Over Time in Diagnostic Tests for Detection of SARS-CoV-2 Infection rrelative to Symptom Onset [13]**

As shown in Figure 1 above, the negative results do not preclude SARS-CoV-2 infection. On top of this, administering RT-PCR tests for the identification of SARS-Cov-2 RNA is complex and needs specifically trained man-power [14]. The SARS-CoV-2 RNA is generally detectable in respiratory specimens during the acute phase of infection. Positive results need to be clinically correlated with patient history and other diagnostic information, after ruling out bacterial infection or co-infection with other viruses. According to Wikramaratna et al. [15], RT-PCR tests are highly specific and the probability of false positives is low but false negatives can also occur if the sample contains insufficient quantities of the virus to be successfully amplified and detected. Xu et al. [16] propose alternative diagnosis via CT scans in combination with deep learning techniques to address relatively low positive rate of RT-PCR Tests in the early stage to determine COVID-19. There is similar work which makes use of X-Rays and CNN [17]. Feng Shi et al. [18] made use of random forest algorithms along with CT-Scans. They claim detection of RT-PCR Tests is only 30-60 percent, requiring repeated tests. As far as other pharmaceutical interventions go, there is work that recommends antibody-based therapies [19] and discovery of potential drugs using deep learning methods [20]. There were studies on effectiveness of hydroxychloroquine and azithromycin, which concluded that for serious patients they did not affect mortality outcomes [21]. More studies may be needed here. K. Kang et al. [22] talk about repurposing existing drugs by making use of AI to predict drug-target interactions. W, Zhang P [23] have studied herd immunity and vaccination game. They inform that herd immunity can be achieved by voluntary, private vaccination. As of now, research is yet to establish long-term post-infection immunity from COVID-19 [24].

Ferguson et al. [25] from Imperial College London classified possible Non-Pharmaceutic Interventions to tackle COVID-19 under mitigation with focus on slowing the epidemic spread and suppression with objective of reversing the epidemic growth. Whereas mitigation included detection, isolation and quarantine of suspect cases, suppression required lock-down of large regions save for bare essentials. They concluded that mitigation is unlikely to be a viable option without overwhelming healthcare systems, suppression is likely necessary in countries able to implement the intensive controls required. Some authors [26] suggested strategies to come out of lock-downs by combining Green Zones and allowing movements in disconnected zones. There has been flurry of response by Mobile App developers to tackle issues related to surveillance with privacy, health and well-being in the context of COVID-19[27]. Yanfang Ye et al. [28] worked on risk assessment of communities by looking at combination of disease-related, demographic, mobility and social media data. There is also research work[29] on monitoring COVID-19 social distancing using person detection.

Kathakali Biswas and Parongama Sen [30] analyzed the space-time dependence of novel Corona Virus outbreak using SIR model and data from China. Their research showed inverse-square law dependence of number of cases against distance from the epicenter. Kathakali Biswas et al. [31] also predicted exponential growth of cases based on SIR model. Their model could also predict how long the disease may last. However. research by Anna L Ziff and Robert M Ziff [32] pointed to exponential growth in death-rates followed by power-law behavior when epidemic peaks and only then exponential decline in death rates. Tim K Tsang et al. [33] did a study on how changing case definitions impacted analysis of disease spread in China, leading to overestimation of basic reproductive numbers. Some studies [34] concluded even after the lockdown of Wuhan on January 23, 2020, the number of patients with serious COVID-19 cases continued to rise, exceeding local hospitalization and ICU capacities for at least a month. Fotios Petropoulos et al. [35] conclude that not only the COVID-19 numbers will grow but also uncertainty about forecasts will also grow. PLOS Editors [36] are vocal in decrying uncertainty in predicting the manifestation of virus in different countries and difficulties in reconciling potentially contradictory data and advice from models and researchers in managing conflicting political, economic and health priorities. Vivek Verma et al. [37] made use of time-to-death periods/lethal periods to predict the morality rates. Vishwesha Guttal [38] in their paper covered risk assessment via layered mobile tracing. Eksin et al. [39] have made modifications to standard SIR/SEIR epidemiological models to account for social distancing to arrive at more realistic predictions.

Some countries have been more successful than others in surveillance of the pandemic. Rapid identification and isolation of cases, quarantine of close contacts, and active monitoring of other contacts have been effective in suppressing expansion of the outbreak in Singapore [40]. South Korea as widely reported in the press, was successful in containing the disease by using anti-body testing. In India however anti-body test setups imported from China failed to work reliably. The paper [41] covered the details of French Surveillance system and early experience with COVID-

19. Kissler et al. [42] have analyzed the potential transmission dynamics through the pandemic period and beyond. They expect the disease to be around at least till 2025 may be with a break.

There are many studies that predicted the growth of COVID-19 cases in India. The very first well-known study [43] was published on ICMR web-site where the focus was on how to handle imported infections. They made use of a variant of SEIR model. Rashmi Pant et al. [44] did a study on cases in Telangana. Dudala et al. [45] applied exponential model for COVID-19 disease progression. Sourish Das [46] applied exponential method derived by SIR model as well as statistical learning model. He estimated that, with lock-down in place, India should in the least get as many cases as in China. He made a study of Basic Reproduction Numbers prevailing in different states in India at the time of his study. The $R_0$, ranged between 2 to 4 for states with national average of 2.75. He particularly called out Punjab for high $R_0$. The Hindu [47] reported that another group of scientists advocated staggered release of lock-downs in India. Tarnjot Kaur et al. [48] emphasized on the importance of paying heed to early warning signals. They projected the number of cases in India to go to 135,000. Ashish Menon et al. [49] studied the cases in India and made use of complex models that increased the number of compartments to many more compared to basic SIR model. They projected huge numbers where deaths are in hundreds of thousands using an overly deterministic model. One of the research opportunities is to make use of social network models, community structures and inherent stochasticity that may temper the overly alarmist projections for COVID-19. Zhaoyang Zhang et al.[50] propose ISIR(Improved SIR) model for epidemics that models spreading of epidemics on Social Contact Networks. This work however pre-dates COVID-19 outbreak.

Many researchers were particularly anguished that COVID-19 accentuated prevailing inequalities further. They [51] make a case that the pandemic presents an opportunity for achieving greater equity in the health care of all vulnerable populations. Similar issues prevail in India where lock-downs are much harder on daily wagers and migrant labourers. This again makes a case for nuanced approaches to tackle epidemics than wholesale lock-downs.

## 3. Proposed Methodology

Hidden Markov Model has great promise in the surveillance of COVID-19 pandemic. This is because certain transitions happening in the progression of virus in the larger society have an association with time series data on confirmed/active cases, recoveries and deaths reported from Hospitals.

A generic Hidden Markov model is illustrated in Figure 2, where the Xi represent the hidden state sequence. The Markov process which is hidden behind the dashed line is determined by the current state and state-transition probability matrix A. We are only able to observe the Oi, which are related to the (hidden) states of the Markov process by the observation probability matrix B. Both the matrices A and B are row-stochastic as well as π which contains initial probabilities of hidden states. A, B and π together constitute the Hidden Markov Model.

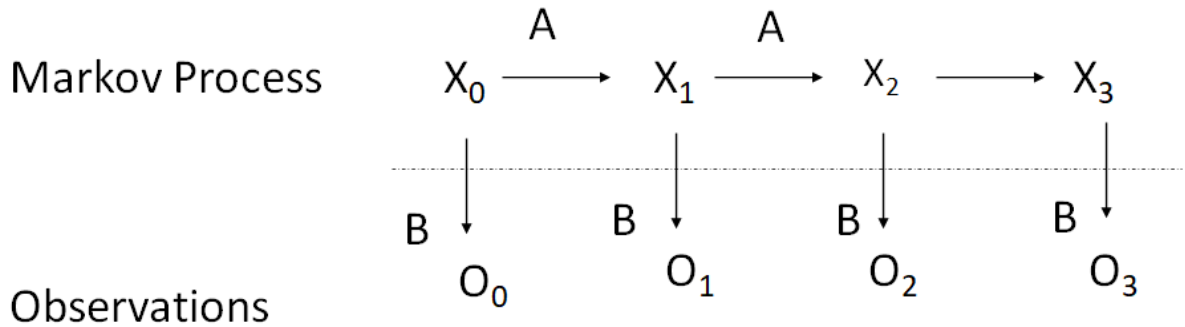

**Figure 2: Hidden Markov Model**

We propose to make use of Hidden Markov Model to perform surveillance of COVID-19 spread. In our model, with each region we associate a state. Thus, we define 4 hidden states which should model majority of situations namely Healthy, Infected, Symptomatic and Detected. In addition, we define 2 additional states that may seldom get used namely Catastrophe-1 and Catastrophe-2. Table 2 below describes the hidden states.

Table 2: States to assess COVID-19 Spread for a given region

| Sr, No. | Name of the State | Description | State Symbol |
|---|---|---|---|
| 1 | **Healthy** | This is the state a region is in prior to the first outbreak of disease and when the region is moving towards normalcy (or non-epidemic phase) | H |
| 2 | **Infected** | Infected people are at different points in incubation phase of 2-14 period. General awareness is low and degree of complacency prevails about the disease due to low cases. | I |
| 3 | **Symptomatic** | This is the state of a region when a region is clearly in epidemic phase and patients come to know about their disease after onset of symptoms, typically in later part of incubation phase. | S |
| 4 | **Detected** | This is the state of a region when region is in epidemic phase and patients come to know about their disease before the onset of symptoms during early part of incubation phase. | D |
| 5 | **Catastrophe-1** | The dynamics of spread suddenly alters due to large number of outward migrations of infected people. | C1 |
| 6 | **Catastrophe-2** | The dynamics of spread suddenly alters due to large number of deaths may be due to health sector capacity issues. | C2 |

The above states are aptly considered hidden as in the absence of exhaustive, expensive and repeated testing it is hard to know what state a given region is in at a given point in time.

Table 3 below represents observations. These observations pertain to reportage on a given day as far as net additions are concerned.

Table 3: Observations based on data reported by Hospitals

| Sr, No. | Name of Category | Description | Remarks | Observation Symbol |
|---|---|---|---|---|
| 1 | **Active** | This observation pertains to additional active cases being reported on a given day/period | These cases are actively being treated by Doctors in Hospitals or tracked remotely | **A** |
| 2 | **Recovered** | This observation pertains to additional recoveries being reported on a given day/period. Total number of recovered cases is generally expected to monotonically increase over time. | A patient is considered recovered if tests on the patient returns negative results. A recovered person may continue to be contagious for some more time or under observation. | **R** |
| 3 | **Dead** | These observations pertain to additional deaths reported on a given day. Total number of deaths reported is generally expected to monotonically increase over time. | In these cases, patients could not recover from the disease and expire during the treatment or brought dead to Hospitals/reported dead to the authorities. | **D** |
| 4 | **Inactive (or Active Complement)** | These observations pertain to reduction in the active number of cases. | Here capacity to treat patients in hospital is freed up due to reduction in active number of cases. | **A-** |
| | **Confirmed** | The cumulative count of above cases is reported under confirmed. | | |

Daily reportage on COVID-19 cases in India is published by the Ministry of Health and Family Welfare, Government of India on their website [52]. India related data is also available in greater detail on Kaggle Platform [53]. For our work, we make use of the former, as it is adequate and convenient. On each day, the public health authorities report aggregate information on confirmed cases, active cases, deaths and recoveries. If we consider each new reportage as an observed symbol comprising of any of

**O** = {Active(**A**), Recovered(**R**), Dead(**D**), Inactive(**A-**)}.

We consider that State/Union Territory goes through the following hidden states:

**S** = {Healthy(**H**), Infected(**I**), Symptomatic(**S**), Detected(**D**)}.

We track 31 States/Union Territories.

This scenario where only Observation Sequences are known, possible state values and possible observations are known lends itself to be modeled as a learning problem. In Learning Problem, given some training observation sequence V and general structure of HMM (number of hidden and visible states), we need to determine HMM parameters **M** = (**A, B, π**) that best fit the training data where A is state-transition probability matrix, B is visible symbol probability matrix and π is probability distribution of hidden states at the initial stage.

For the COVID-19 Surveillance scenario, we propose the following **alternative solution approach** to learning problem instead of standard Baum-Welch forward-backward EM algorithm.

1. Choose a vector of observation sequences catering to different region. Each region on a given day can generate a, d, r and a- symbols in different numbers that account for change of observed states of patients on a given day.
2. Based on the visible symbols generated above we infer state symbols for that day as follows.

Table 4: Inferring Hidden States from Daily Observations

| State | Active (Net Additions) | Recovered (Additions) | Dead (Additions) |
|---|---|---|---|
| **Healthy** | Low | High | Medium |
| **Infected** | High | Low | Medium |
| **Symptomatic** | High | Medium | Low |
| **Detected** | Medium | High | Low |
| **Catastrophe-1** | Low | Medium | High |
| **Catastrophe-2** | Medium | Low | High |

3. For each region and each day generate the π, the probability distribution of hidden states at that period.
4. For each day and for each hidden state value compute the visible symbol distribution matrix and compute/update visible symbol probability distribution matrix.
5. Compare the state inferred for the previous day and current day for different regions and populate the state-transition distribution matrix. Update the State Transition Probability matrix with incremental state transitions for every day. This starts getting updated from day 2.
6. Iterate between step 1 to 4 till symbols are exhausted.

Thus, at the end of steps 1 to 6, we have a Hidden Markov Model comprising of A, B and π. The model can then be used for Evaluation Problem and Decoding Problem. The state distributions arrived using this model on a given day can be used by Governments to design interventions in a calibrated manner. We can further refine the model by using numeric values for probability for B from the model in step 2 and arrive at more refined values of A.

## 4. Results

We made use of the data on COVID-19 published on daily basis by the Ministry of Health and Family Welfare [52], Government of India to train the HMM. The approach we chose was to make reasonable assumptions about mapping between data reported by Hospitals and state of infection in regions as shown in Table 4. Then by making use of inferred states we derive state transitions. We followed this approach for data between 29$^{th}$ April and May 8$^{th}$, which is detailed in Tables 7A and 7B. Then based on the derived state transitions we arrive at State Transition Matrix(A). The computation of which is described in Table 8. Now based on our knowledge of States inferred, we compute the observation probability Matrix B as shown in Table 9. Thus, based on Analysis of data between April 29$^{th}$ and May 8$^{th}$ we have a refined Model parameter B with quantitative values compared to qualitative inferences we started with. We also have learnt the HMM parameter A i.e State Transition Matrix. Now the model learnt is described in Table 10. To enable regional analysis, we provide Table 11 that lists the States/UTs based on their Hidden States and Table 12 that reports hidden state sequences for last one week for each of the states and UTs.

We have chosen to use the model in Table 10 only for prospective data (and not retrospectively). We believe this is an appropriate decision as we are primarily looking at sequential data

Figure 3 illustrates state transitions and Figure 4 depicts corresponding mapping between observations and inferred states. Figure 5 shows the distribution of inferred states. In all these cases we have used our initial approach for data between April 29$^{th}$ and May 8$^{th}$ and then use the learnt model for data between May 9$^{th}$ to 20$^{th}$

The approach we recommend is to recompute and refine the HMM every two weeks and apply for the prospective data.

We made use of the refined Model as of May 20$^{th}$ for the period May 21 to 25 and data as of June 10$^{th}$. The Hidden State Sequences region-wise along with active cases all the way to June 10, 2020 are shown in Table 14.

Figures 6 to Figure 9 illustrate hidden state transition probabilities of states(regions) of concern compared to similar states(regions). Here we assume that the transition sequence leads us towards normalization and hence benign.

**Healthy(H)->Infected(I)->Symptomatic(S)->Detected(D)->Healthy(H)**

These transitions are represented clock-wise and with solid-lines. The transitions in opposite sequence are considered as matter of concern and they are represented using dashed-lines and/or in anti-clockwise manner. Transition back to the same state is considered benign in case of Healthy States as well as Detected States. The latter ensures timely detection of pandemic spread. Transition back to the same state is considered a matter of concern in case of Symptomatic as well as Infected states. As earlier benign transitions are marked using solid lines and others using dashd lines.

Further after a gap of a month we reviewed the data on COVID-19 spread in India on July 11$^{th}$ and arrived at the hidden states with June 10$^{th}$ as the base date. The status of states as on July 11$^{th}$ is given in Table 15. Then again, we assessed the incremental spread on July 12$^{th}$ relative to data on the 11$^{th}$ and the inferred hidden states of different states and union territories are given in Table 16.

## 5. Recommendations

In this section we analyse the results we have obtained in order to make policy/process recommendations. We can make the following observations regarding COVID-19 Pandemic Surveillance in India. It is to be noted that because of inter-state movements of migrants there was some impact on the COVID-19 scenario where certain states may have reported surge in the number of cases due to influx of migrants. Here Figure 5, Table 11 and Table 12 illustrate the data as of May 20$^{th}$ and Figure 6-9 illustrate the data as of June 10$^{th}$

1. As shown in Figure 5, the number of healthy states has remained around the same and new infections seem to be still breaking out. On the positive side there seems to be steady reduction in the symptomatic states and increase in the number of detected states. This means there is early detection of the cases to greater degree than earlier.
2. Table 11 which shows data as on May 20$^{th}$, indicates that there are Healthy States such as Punjab and Andhra Pradesh while they continue to have high cases in Hospital they are moving towards normalcy. Then we have Infected states such as Assam, Goa and Karnataka which have relatively smaller number of cases in the Hospital but continue to have fresh infections. Then we have Symptomatic States namely Uttar Pradesh, Rajasthan, Telangana and Bihar that continue to report new cases. Then we have nine other states such as West Bengal which have made improvement as far as detection of cases is concerned.
3. Table 12, which gives the hidden state sequence for last one week gives greater clarity. Here Uttar Pradesh is predominantly in Detected State, once in a while moving to Symptomatic and Infected. Andhra Pradesh and Punjab are indeed moving towards normalcy. Only some UTs and smaller states are able to maintain the Healthy State. Rest of the states continue to be in flux as far as coming to normalcy is concerned. They still need to do work in controlling new infections. Here Gujarat which has seen highest mortalities is consistently showing Detected state. This indicates more work is required on Pharmaceutical side to take care of possibly more vulnerable patients due to comorbidities. The same can be said about West Bengal. On the other hand, Maharashtra which has very large number of cases showing Symptomatic State to greater measure, indicating more work required to detect cases early. In many other states, situation is yet to stabilize.
4. Figure 6 compares Uttara Pradesh with 4,365 active cases and Maharashtra with 44,860 active cases. Here we can clearly see that in case of Maharashtra, 91 percent of transitions are from symptomatic state back to itself with no transition to healthy or infected state whatsoever. This indicates huge localized spread which may be due to cases in Mumbai with exceptionally high population density. In case of Uttar Pradesh which is most populous state in India 57 percent of the time Healthy state reverts back to itself. This however needs to be balanced with the observation that symptomatic state also returns back to itself. On the positive side there are transitions between all the states giving an indication that situation may normalize sooner than later.

5. Figure 7 compares state of Delhi with neighbouring state of Haryana. Delhi has 18,453 cases and Haryana has 3,030. One can see that there continue to be outbreak of infections in Delhi whereas Haryana never goes to infected state. Delhi in 73 percent of the cases goes back to symptomatic state. There are direct transitions between infected and detected states, indicating availability of adequate testing infrastructure. Delhi also stays 60 percent of the time in Detected state further reinforcing the same observation. Overall while things continue to be of concern in Delhi, they most probably will normalize much sooner than Maharashtra provided new outbreaks of infection are prevented. In case of Haryana we can broadly say that 40 percent of the cases it is in healthy state and 60 percent of the cases it is in symptomatic state. Here the focus should be on active cases and how to get them cured fast.

6. Figure 8 compares Gujarat and West Bengal which had high death rates compared to other regions. As of June 10$^{th}$, Gujarat had 5,336 active cases and West Bengal had 4,959 active cases. In case of Gujarat there is no transition to infected state indicating success in separation of infected zones from other zones. In case of West Bengal there continued be transitions to infected zones. This may call for more action to enforce social distancing and containment. In other respects, both the states seem to behave similarly.

7. Figure 9 compares state of Tamil Nadu with 16282 active cases with the neighbouring state of Karnataka which has 3,251. Both states have similar population sizes and densities. Tamil Nadu has no transition back into Healthy State whereas Karnataka is in Healthy State 57 percent of time. Karnataka has no activity in Infected State whereas Tamil Nadu has. In other respects, profile of Tamil Nadu represents that of Delhi, which is discussed earlier. One reason Karnataka to did better is the seriousness with which they administered lock-down as well as the way they involved Medical Experts [54].

8. Table 15 represents inferred states as of July 11$^{th}$ using the data as of June 10$^{th}$ as the base. Thus, these states may be considered to represent long term trend. Within a month around 25 out of 31 states are in detected state, reflecting that there is much greater proportion of cases are detected early. Thus, greater focus is required on improving the health-care infrastructure than restrictions such as lock-downs. Five states are showing up as infected. Among them Karnataka has announced[55] that in couple of districts in and around Bengaluru they will be imposing a lock-down between July 14$^{th}$ and 22$^{nd}$. Only one state is showing up as healthy indicating the continued challenges country as a whole face. Overall, we can say situation on the ground corroborates with our findings.

9. Table 16 represents the inferred states reflecting incremental spread of epidemic on July 12$^{th}$ relative to day before. This shows 4 states including Delhi under healthy category, indicating that these states are moving towards normalcy. Among others, 5 states are showing up as infected, 8 symptomatic and 14 detected, thereby calling for appropriate day-to-day tactical interventions.

In summary, we make the following recommendations.

- States/UTs/zones which predominantly maintain Healthy State, should strictly regulate migrations in and out of the region.

- For States/UTs/zones which show off and on, infected status, focus should be on those who may potentially get infected. These may be health workers or people who typically interact with large number of people or move around a lot.
- For States/UTs/zones which maintain symptomatic status, the focus should be on non-pharmaceutical interventions such as social distance and containment.
- State/UTs which maintain detected status yet may have high mortality the focus should be on pharmaceutical interventions and on improving the health infrastructure.

**Table 7A: COVID-19 Progression in India between 29-4-2020 and 3-5-2020**

| | State | 29/04/2020 | | | | 30/04/2020 | | | | 01/05/2020 | | | | 02/05/2020 | | | | 03/05/2020 | | | |
|---|---|---|---|---|---|---|---|---|---|---|---|---|---|---|---|---|---|---|---|---|---|
| | | A | R | D | State Symbol | A | R | D | State Symbol | A | R | D | State Symbol | A | R | D | State Symbol | A | R | D | State Symbol |
| 1 | Uttar Pradesh | -10 | 77 | 5 | H | 49 | 36 | 3 | S | 34 | 42 | 2 | S | 71 | 101 | 2 | D | 129 | 42 | 0 | S |
| 2 | Andhrra Pradesh | 44 | 29 | 0 | S | 37 | 34 | 0 | S | -24 | 82 | 2 | H | 24 | 38 | 0 | S | 11 | 47 | 0 | D |
| 3 | Punjab | 8 | 0 | 1 | I | 16 | 19 | 0 | D | 0 | 0 | 0 | D | 392 | 22 | 1 | S | 0 | 0 | 0 | S |
| 4 | Chandigarh | 16 | 0 | 0 | S | 0 | 0 | 0 | S | 0 | 0 | 0 | S | 32 | 0 | 0 | S | 4 | 2 | 0 | S |
| 5 | Maharashtra | 591 | 106 | 31 | S | 360 | 205 | 32 | S | 376 | 180 | 27 | S | 876 | 106 | 26 | S | 633 | 121 | 36 | S |
| 6 | Telengana | -73 | 87 | 0 | H | 0 | 0 | 0 | H | -47 | 74 | 0 | H | 18 | 0 | 0 | S | -13 | 17 | 2 | H |
| 7 | Jharkhand | 0 | 2 | 0 | D | 2 | 0 | 0 | D | 3 | 1 | 0 | S | 0 | 0 | 0 | S | 2 | 2 | 0 | S |
| 8 | Ladakh | 0 | 0 | 0 | S | 0 | 0 | 0 | H | 0 | 0 | 0 | H | -1 | 1 | 0 | H | 0 | 0 | 0 | S |
| 9 | Gujarat | 137 | 40 | 19 | S | 229 | 93 | 16 | S | 210 | 86 | 17 | S | 182 | 122 | 22 | S | 147 | 161 | 26 | D |
| 10 | Tamilnadu | 53 | 1058 | 1 | D | 60 | 42 | 2 | S | 113 | 48 | 0 | S | 148 | 54 | 1 | S | 189 | 35 | 1 | S |
| 11 | Uttarkhanda | 2 | 1 | 0 | S | -1 | 2 | 0 | D | 2 | 0 | 0 | S | 0 | 1 | 0 | D | -1 | 2 | 0 | H |
| 12 | Meghalaya | 0 | 0 | 0 | S | 0 | 0 | 0 | S | 0 | 0 | 0 | S | 0 | 0 | 0 | S | 0 | 0 | 0 | S |
| 13 | West Bengal | 16 | 10 | 2 | S | 28 | 5 | 0 | S | 11 | 15 | 11 | D | 0 | 0 | 0 | D | 115 | 12 | 0 | S |
| 14 | Himachal | -3 | 3 | 0 | H | 0 | 0 | 0 | H | -3 | 3 | 0 | H | -2 | 2 | 0 | H | -3 | 3 | 0 | H |
| 15 | Goa | 0 | 0 | 0 | H | 0 | 0 | 0 | H | 0 | 0 | 0 | H | 0 | 0 | 0 | H | 0 | 0 | 0 | H |
| 16 | Rajastan | -81 | 251 | 10 | H | 72 | -1 | 2 | H | 73 | 68 | 5 | S | -202 | 280 | 4 | H | 98 | 5 | 3 | S |
| 17 | Jammu and Kashmir | 5 | 12 | 1 | D | 0 | 16 | 0 | D | 9 | 24 | 0 | D | -6 | 31 | 0 | H | 20 | 7 | 0 | D |
| 18 | Assam | 0 | 0 | 0 | S | 2 | 2 | 0 | S | 0 | 0 | 0 | S | -2 | 3 | 0 | H | 0 | 0 | 0 | H |
| 19 | Puducherry | 0 | 0 | 0 | S | -2 | 2 | 0 | H | 0 | 0 | 0 | H | 0 | 0 | 0 | H | 0 | 0 | 0 | H |
| 20 | Madhya Pradesh | 171 | 11 | 6 | S | 4 | 89 | 11 | H | 31 | 21 | 7 | S | -50 | 42 | 8 | H | -158 | 274 | 11 | H |
| 21 | Karnataka | -14 | 17 | 0 | H | 25 | 8 | 1 | S | 6 | 12 | 1 | D | -1 | 20 | 3 | H | -19 | 27 | 0 | H |
| 22 | Chattisghar | -1 | 2 | 0 | H | -2 | 2 | 0 | H | 2 | 0 | 0 | H | 3 | 0 | 0 | S | 0 | 0 | 0 | S |
| 23 | Tripura | 0 | 0 | 0 | S | 0 | 0 | 0 | H | 0 | 0 | 0 | H | 0 | 0 | 0 | H | 2 | 0 | 0 | H |
| 24 | Kerala | 1 | -7 | 0 | S | 0 | 10 | 0 | D | -13 | 14 | 0 | H | -8 | 9 | 0 | H | -6 | 8 | 0 | H |
| 25 | Andaman | -4 | 4 | 0 | H | 0 | 0 | 0 | S | -1 | 1 | 0 | H | 0 | 0 | 0 | H | 0 | 0 | 0 | H |
| 26 | Bihar | 31 | 7 | 0 | S | 19 | 1 | 0 | S | 6 | 17 | 0 | D | 28 | 16 | 1 | S | -9 | 19 | 1 | H |
| 27 | Arunachal | 0 | 0 | 0 | S | 0 | 0 | 0 | H | 0 | 0 | 0 | H | 0 | 0 | 0 | H | 0 | 0 | 0 | H |
| 28 | Haryana | -12 | 26 | 0 | H | 0 | 0 | 0 | H | 3 | 0 | 0 | H | 28 | 18 | 1 | H | 34 | 0 | 0 | S |
| 29 | Manipur | 0 | 0 | 0 | S | 0 | 0 | 0 | H | 0 | 0 | 0 | H | 0 | 0 | 0 | H | 0 | 0 | 0 | H |
| 30 | Delhi | 5 | 201 | 0 | S | 109 | 14 | 2 | S | 71 | 2 | 3 | S | 148 | 73 | 2 | S | 292 | 89 | 3 | S |
| 31 | Odisha | 0 | 1 | 0 | S | 18 | 1 | 0 | S | 3 | 2 | 0 | S | -3 | 14 | 0 | H | 5 | 1 | 0 | S |

**Table 7B: COVID-19 Progression in India between 4-5-2020 and 8-5-2020**

| | State | 04/05/2020 | | | | 05/05/2020 | | | | 06/05/2020 | | | | 07/05/2020 | | | | 08/05/2020 | | | |
|---|---|---|---|---|---|---|---|---|---|---|---|---|---|---|---|---|---|---|---|---|---|
| | | A | R | D | State Symbol | A | R | D | State Symbol | A | R | D | State Symbol | A | R | D | State Symbol | A | R | D | State Symbol |
| 1 | Uttar Pradesh | 54 | 60 | 2 | D | -77 | 186 | 8 | H | -25 | 43 | 3 | H | -29 | 143 | 4 | H | -50 | 121 | 2 | H |
| 2 | Andhrra Pradesh | 28 | 36 | 3 | D | 2 | 65 | 0 | D | 0 | 0 | 0 | D | -80 | 140 | 0 | H | 17 | 51 | 2 | H |
| 3 | Punjab | 324 | 5 | 1 | S | 118 | 11 | 2 | S | 211 | 5 | 2 | S | 61 | 2 | 2 | S | 113 | 14 | 1 | S |
| 4 | Chandigarh | 0 | 0 | 0 | S | 5 | 2 | 1 | S | 9 | 0 | 0 | S | 9 | 0 | 0 | S | 15 | 0 | 0 | S |
| 5 | Maharashtra | 536 | 115 | 27 | S | 1182 | 350 | 35 | S | 596 | 354 | 34 | S | 924 | 275 | 34 | S | 966 | 207 | 43 | S |
| 6 | Telengana | -14 | 32 | 1 | H | -92 | 95 | 0 | H | 11 | 0 | 0 | S | -32 | 43 | 0 | H | -6 | 22 | 0 | H |
| 7 | Jharkhand | 0 | 0 | 0 | S | -5 | 5 | 0 | H | 4 | 6 | 0 | D | -2 | 4 | 0 | H | 1 | 4 | 0 | D |
| 8 | Ladakh | 1 | 0 | 0 | S | 0 | 0 | 0 | S | 0 | 0 | 0 | S | 0 | 0 | 0 | S | 1 | 0 | 0 | S |
| 9 | Gujarat | 199 | 146 | 28 | S | 194 | 153 | 29 | S | 206 | 186 | 49 | S | 233 | 119 | 28 | S | 149 | 209 | 29 | D |
| 10 | Tamilnadu | 239 | 32 | 1 | S | 496 | 30 | 1 | S | 430 | 76 | 2 | S | 738 | 31 | 2 | S | 547 | 31 | 2 | S |
| 11 | Uttarkhanda | 0 | 0 | 1 | H | 0 | 0 | 0 | H | 1 | 0 | 0 | S | 0 | 0 | 0 | S | 0 | 0 | 0 | S |
| 12 | Meghalaya | 0 | 0 | 0 | S | -10 | 10 | 0 | S | 0 | 0 | 0 | S | 0 | 0 | 0 | S | 0 | 0 | 0 | S |
| 13 | West Bengal | 39 | 0 | 2 | I | 131 | 67 | 98 | S | -68 | 146 | 7 | H | 108 | 0 | 4 | S | 85 | 0 | 7 | I |
| 14 | Himachal | -1 | 1 | 0 | H | -3 | 4 | 0 | H | 0 | 0 | 1 | H | 3 | 0 | 0 | S | 1 | 0 | 0 | S |
| 15 | Goa | 0 | 0 | 0 | H | 0 | 0 | 0 | H | 0 | 0 | 0 | H | 0 | 0 | 0 | H | 0 | 0 | 0 | H |
| 16 | Rajastan | -127 | 235 | 6 | H | 131 | 38 | 6 | S | -46 | 131 | 12 | H | 85 | 71 | 3 | S | 105 | 0 | 5 | I |
| 17 | Jammu and Kashmir | 2 | 33 | 0 | D | 9 | 16 | 0 | D | -2 | 17 | 0 | H | 32 | 2 | 0 | D | 4 | 13 | 1 | D |
| 18 | Assam | 0 | 0 | 0 | H | 0 | 0 | 0 | H | 0 | 0 | 0 | H | 2 | 0 | 0 | D | 7 | 2 | 0 | S |
| 19 | Puducherry | 0 | 0 | 0 | H | 0 | 1 | 0 | D | 0 | 0 | 0 | D | 0 | 0 | 0 | D | 0 | 0 | 0 | H |
| 20 | Madhya Pradesh | 87 | 0 | 9 | S | -106 | 202 | 11 | H | 0 | 0 | 0 | H | -19 | 99 | 9 | H | -26 | 132 | 8 | H |
| 21 | Karnataka | 13 | 22 | 1 | D | -5 | 20 | 2 | H | 4 | 7 | 1 | H | -1 | 23 | 0 | H | -1 | 12 | 1 | H |
| 22 | Chattisghar | 14 | 0 | 0 | S | 1 | 0 | 0 | H | 1 | 0 | 0 | S | 0 | 0 | 0 | S | -2 | 2 | 0 | H |
| 23 | Tripura | 12 | 0 | 0 | S | 13 | 0 | 0 | S | 14 | 0 | 0 | S | 0 | 0 | 0 | S | 22 | 0 | 0 | S |
| 24 | Kerala | -1 | 1 | 0 | H | -61 | 61 | 0 | H | 2 | 0 | 0 | S | -6 | 7 | 0 | S | -5 | 5 | 0 | H |
| 25 | Andaman | -15 | 15 | 0 | H | 0 | 0 | 0 | H | 0 | 0 | 0 | H | 0 | 0 | 0 | H | -1 | 1 | 0 | H |
| 26 | Bihar | 27 | 8 | 0 | S | 7 | 5 | 0 | S | -5 | 12 | 0 | H | -40 | 46 | 0 | H | -51 | 58 | 1 | H |
| 27 | Arunachal | 0 | 0 | 0 | H | 0 | 0 | 0 | H | 0 | 0 | 0 | H | 0 | 0 | 0 | H | 0 | 0 | 0 | H |
| 28 | Haryana | 30 | 18 | 0 | S | 64 | 9 | 2 | S | -33 | 2 | 62 | H | 41 | 4 | 1 | S | 31 | 0 | 0 | S |
| 29 | Manipur | 0 | 0 | 0 | H | 0 | 0 | 0 | H | 0 | 0 | 0 | H | 0 | 0 | 0 | H | 0 | 0 | 0 | H |
| 30 | Delhi | 321 | 106 | 0 | S | 280 | 69 | 0 | S | 169 | 37 | 0 | S | 353 | 74 | 1 | S | 58 | 389 | 1 | D |
| 31 | Odisha | -1 | 4 | 0 | H | 7 | 0 | 0 | S | 5 | 0 | 0 | S | 8 | 1 | 1 | S | 33 | 1 | 0 | S |

### Table 8: Computation of HMM Parameter A State Transition Probability Matrix

| | Source | Target states | | | | Source | | | | | Source | Target States | | | | Source | Target States | | | |
|---|---|---|---|---|---|---|---|---|---|---|---|---|---|---|---|---|---|---|---|---|
| Date | | H | I | S | D | | H | I | S | D | | H | I | S | D | | H | I | S | D |
| 30/4/2020 | Healthy | 10 | | 6 | | Infected | | | | 1 | Symptoma | 2 | | 8 | 2 | Detcted | | | | 3 |
| 1/5/2020 | | 10 | | 3 | | | | | | | | 2 | | 9 | 3 | | 1 | | 2 | 2 |
| 2/5/2020 | | 10 | | 3 | | | | | | | | 4 | | 8 | 2 | | 2 | | 2 | 1 |
| 3/5/2020 | | 11 | | 4 | 1 | | | | | | | 2 | | 9 | 2 | | 1 | | 1 | |
| 4/5/2020 | | 10 | | 3 | 1 | | | | | | | 2 | 1 | 11 | 1 | | | | 1 | 2 |
| 5/5/2020 | | 10 | | 2 | 1 | | | | 1 | | | 3 | | 10 | | | 2 | | | 2 |
| 6/5/2020 | | 9 | | 4 | 1 | | | | | | | 4 | | 11 | | | 1 | | | 2 |
| 7/5/2020 | | 8 | | 4 | 2 | | | | | | | 1 | | 14 | | | 2 | | | 1 |
| 8/5/2020 | | 10 | | | 1 | | | | | | | 2 | 2 | 11 | 2 | | 1 | | 1 | 1 |
| Total | 124 | 88 | 0 | 29 | 7 | | 2 | 0 | 0 | 1 | 1 | 128 | 22 | 3 | 91 | 12 | 31 | 10 | 0 | 7 | 14 |
| | | 0.710 | 0 | 0.234 | 0.056 | | 0.000 | 0.000 | 0.5 | 0.5 | | 0.172 | 0.023 | 0.711 | 0.094 | | 0.323 | 0 | 0.226 | 0.452 |

### Table 9: Computation of HMM Parameter B Observation Probability Matrix

| | Healthy | | | Infected | | | Symptomatic | | | Detected | | |
|---|---|---|---|---|---|---|---|---|---|---|---|---|
| Date | Active | Recovered | Dead | Active | Recovered | Dead | Active | Recovered | Dead | Active | Recovered | Dead |
| 29/4/2020 | -188 | 467 | 15 | 8 | 0 | 1 | 1009 | 406 | 58 | 58 | 1072 | 2 |
| 30/4/2020 | 68 | 94 | 13 | | | | 936 | 441 | 56 | 19 | 47 | 0 |
| 1/5/2020 | -80 | 171 | 2 | | | | 916 | 450 | 61 | 32 | 68 | 12 |
| 2/5/2020 | -246 | 411 | 16 | | | | 1851 | 437 | 53 | 71 | 102 | 2 |
| 3/5/2020 | -203 | 350 | 14 | | | | 1411 | 409 | 43 | 178 | 215 | 26 |
| 4/5/2020 | -159 | 288 | 8 | 39 | 0 | 2 | 1790 | 422 | 66 | 97 | 151 | 6 |
| 5/5/2020 | -348 | 573 | 21 | | | | 1208 | 744 | 174 | 11 | 82 | 0 |
| 6/5/2020 | -169 | 358 | 88 | | | | 1655 | 358 | 86 | 4 | 6 | 0 |
| 7/5/2020 | -203 | 488 | 13 | | | | 2557 | 584 | 76 | 34 | 2 | 0 |
| 8/5/2020 | -125 | 404 | 14 | 190 | 0 | 12 | 1736 | 255 | 46 | 212 | 615 | 31 |
| Total | -1653 | 3604 | 204 | 237 | 0 | 15 | 15069 | 4506 | 719 | 716 | 2360 | 79 |
| Probabilties | | | | | | | | | | | | |
| Sign changed | 1653 | 3604 | 204 | 237 | 0 | 15 | 15069 | 4506 | 719 | 716 | 2360 | 79 |
| A- | A | R | | D | A | R | D | A | R | D | A | R | D |
| 0.303 | NA | 0.65995 | 0.03736 | 0.94048 | 0.00000 | 0.05952 | 0.74253 | 0.22204 | 0.03543 | 0.22694 | 0.74802 | 0.02504 |
| 30.269 | | 65.995 | 3.736 | 94.048 | 0.000 | 5.952 | 74.253 | 22.204 | 3.543 | 22.694 | 74.802 | 2.504 |

### Table 10: Hidden Markov Model for COVID-19 Surveillance (A, B)

| | Healthy | Infected | Symptomatic | Detected |
|---|---|---|---|---|
| Healthy | 0.710 | 0 | 0.234 | 0.056 |
| Infected | 0.000 | 0.000 | 0.5 | 0.5 |
| Symptomatic | 0.172 | 0.023 | 0.711 | 0.094 |
| Detected | 0.323 | 0 | 0.226 | 0.452 |

| | Active (-) | Active | Recovered | Dead |
|---|---|---|---|---|
| Healthy | 0.303 | 0 | 0.65995 | 0.03736 |
| Infected | | 0.94048 | 0.00000 | 0.05952 |
| Symptomatic | | 0.74253 | 0.22204 | 0.03543 |
| Detected | | 0.22694 | 0.74802 | 0.02504 |

**Table 11 Status of States and UTs on May 20, 2020**

| Healthy | Infected | Symptomatic | Detected |
|---|---|---|---|
| Andhra Pradesh | Uttarakhand | Uttar Pradesh | Chandigarh |
| Punjab | Goa | Telengana | Maharashtra |
| Jharkhand | Assam | Rajastan | Gujarat |
| Ladakh | Karnataka | Bihar | TN |
| Meghalaya | Chhattisgarh | | WB |
| Himachal | Kerala | | MP |
| J&K | Manipur | | Haryana |
| Puducherry | | | Delhi |
| Tripura | | | Odisha |
| Andaman | | | |
| Arunachal | | | |

**Table 12 Hidden State Sequences for States and Union Territories**

| | 14-5-2020 | 15-5-2020 | 16-5-2020 | 17-5-2020 | 18-5-2020 | 19-5-2020 | 20-05-2020 |
|---|---|---|---|---|---|---|---|
| Uttar Pradesh | S | D | D | D | I | D | S |
| Andhra Pradesh | H | D | D | D | H | H | H |
| Punjab | H | H | H | H | H | H | H |
| Chandigarh | H | H | H | H | H | D | D |
| Maharashtra | S | S | S | S | S | S | D |
| Telengana | H | S | S | S | D | S | S |
| Jharkhand | I | S | I | I | I | H | H |
| Ladakh | D | H | H | H | H | H | H |
| Gujarat | D | D | D | D | D | D | D |
| Tamilnadu | I | S | D | D | D | S | D |
| Uttarakhand | I | D | S | S | S | I | I |
| Meghalaya | H | H | H | H | H | H | H |
| West Bengal | D | D | D | D | D | S | D |
| Himachal | I | I | S | S | D | I | H |
| Goa | H | I | H | H | I | I | I |
| Rajastan | S | D | D | D | D | D | S |
| Jammu and Kashmir | S | H | D | D | D | S | H |
| Assam | I | I | D | D | I | I | I |
| Puducherry | H | H | H | H | H | I | H |
| Madhya Pradesh | D | D | D | D | S | I | D |
| Karnataka | D | S | S | S | S | S | I |
| Chattisghar | H | D | I | I | S | I | I |
| Tripura | H | H | H | H | H | H | H |
| Kerala | I | I | I | I | S | I | I |
| Andaman | H | H | H | H | H | H | H |
| Bihar | I | S | H | H | S | S | S |
| Arunachal | H | H | H | H | H | H | H |
| Haryana | H | D | H | H | H | H | D |
| Manipur | H | I | H | H | H | H | I |
| Delhi | D | S | H | H | D | H | D |
| Odisha | S | S | S | S | D | I | D |

**Table 13 Refined Hidden Markov Model for COVID-19 Surveillance (A, B) to be used from May 21ˢᵗ onwards**

|  | Healthy | Infected | Symptomatic | Detected |
|---|---|---|---|---|
| **Healthy** | 0.668 | 0.04 | 0.166 | 0.130 |
| **Infected** | 0.193 | 0.439 | 0.19 | 0.18 |
| **Symptomatic** | 0.162 | 0.09 | 0.624 | 0.124 |
| **Detected** | 0.260 | 0.09 | 0.260 | 0.39 |

|  | Active (-) | Active | Recovered | Dead |
|---|---|---|---|---|
| **Healthy** | 0.323 | 0.005 | 0.653 | 0.020 |
| **Infected** | 0.000 | 0.901 | 0.083 | 0.017 |
| **Symptomatic** | 0.000 | 0.708 | 0.262 | 0.031 |
| **Detected** | 0.003 | 0.249 | 0.714 | 0.034 |

**Table 14: State-wise Hidden State Sequences between 29-4-2020 and 10-June 2020**

| # | State | Active cases on June 10 | 29-Apr | 30-Apr | 1-May | 2-May | 3-May | 4-May | 5-May | 6-May | 7-May | 8-May | 9-May | 10-May | 11-May | 13-May | 14-May | 15-May | 16-May | 17-May | 18-May | 19-May | 20-May | 22-May | 23-May | 24-May | 25-May | 10-Jun |
|---|---|---|---|---|---|---|---|---|---|---|---|---|---|---|---|---|---|---|---|---|---|---|---|---|---|---|---|---|
| 1 | Uttar Pradesh | 4365 | H | S | S | D | S | D | H | H | H | H | H | D | D | H | S | D | D | H | I | D | S | S | S | D | D | D |
| 2 | AP | 2191 | S | S | H | S | D | D | D | D | H | H | H | D | D | H | H | D | D | H | H | H | D | D | D | D | S |
| 3 | Punjab | 497 | I | D | D | S | S | S | S | S | S | S | S | I | I | H | H | H | H | H | H | H | H | H | H | S |
| 4 | Chandigarh | 32 | S | S | S | S | S | S | S | S | S | S | S | S | H | H | H | H | H | H | D | D | H | H | I | D | H |
| 5 | Maharshtra | 44860 | S | S | S | S | S | S | S | S | S | S | S | S | S | S | S | S | S | D | S | S | S | S | S | D |
| 6 | Telangana | 1963 | H | H | H | S | H | H | H | S | H | H | H | I | D | H | S | S | S | D | S | S | S | I | D | D | S |
| 7 | Jharkhand | 844 | D | D | S | S | S | S | H | D | H | D | H | D | I | I | S | I | H | I | H | I | S | I | S | S |
| 8 | Ladakh | 55 | H | H | H | H | S | S | S | S | S | S | H | H | D | H | H | H | H | H | I | H | I | I | S |
| 9 | Gujarat | 5336 | S | S | S | S | D | S | S | S | S | D | S | D | H | D | D | D | S | D | D | D | D | H | D | D | H |
| 10 | Tamil Nadu | 16282 | D | S | S | S | S | S | S | S | S | S | S | S | I | S | D | H | D | S | D | D | D | S | D | D |
| 11 | Uttarakhand | 769 | S | D | S | D | H | C1(H) | H | S | S | S | H | S | I | I | D | S | I | S | I | I | I | S | I | I | D |
| 12 | Meghalaya | 29 | S | S | S | S | S | S | S | S | S | I | H | H | H | H | H | H | H | H | I | H | H | H |
| 13 | West Bengal | 4959 | S | S | D | D | S | I | S | H | S | I | S | I | S | D | D | D | S | D | S | D | D | D | I | S | S |
| 14 | Himachal | 191 | H | H | H | H | H | H | C1(H) | S | S | H | I | I | I | I | S | H | D | I | H | S | I | I | I | D |
| 15 | Goa | 292 | H | H | H | H | H | H | H | H | H | H | H | H | H | H | I | H | I | I | I | I | H | I | I | S |
| 16 | Rajastan | 2662 | H | H | S | H | S | H | S | H | S | I | H | D | H | D | S | D | D | D | D | S | S | D | S | S | D |
| 17 | Jammu and Kashmir | 2792 | D | D | D | H | D | D | D | H | D | D | H | S | D | D | S | H | D | S | D | S | H | S | D | D | D |
| 18 | Assam | 1848 | H | S | S | H | H | H | H | H | D | S | S | I | H | H | I | I | D | I | I | I | I | S | I | I | S |
| 19 | Puducherry | 75 | S | H | H | H | H | H | D | D | D | H | H | H | D | H | H | H | H | I | H | S | I | H | I | S |
| 20 | MP | 2700 | S | H | S | H | S | H | H | H | H | H | H | H | D | D | D | S | S | I | D | S | H | D | S | D |
| 21 | Karnataka | 3251 | H | S | D | H | D | H | H | S | H | H | S | D | I | D | S | S | S | S | I | I | S | I | I | S |
| 22 | Chattisgarh | 848 | H | H | H | S | S | S | H | S | H | H | H | H | D | I | S | I | I | I | I | I | S |
| 23 | Tripura | 671 | H | H | H | H | H | S | S | S | S | S | I | I | H | H | H | H | H | H | I | H | I |
| 24 | Kerala | 1232 | S | D | H | H | H | S | H | S | H | D | I | S | I | I | S | I | I | S | I | I | I | S |
| 25 | Andaman &Nicobar | 0 | H | S | H | H | H | H | H | H | H | H | H | H | H | H | H | H | H | H | H | H | H | H | H |
| 26 | Bhar | 2563 | H | S | D | S | H | S | S | H | H | H | H | S | S | I | S | H | S | S | S | I | I | S | D |
| 27 | Arunachala | 56 | H | H | H | H | H | H | H | H | H | H | H | H | H | H | H | H | H | H | H | H | H | I |
| 28 | Haryana | 3030 | H | H | H | H | S | S | S | H | S | H | S | D | H | D | H | D | H | H | D | D | D | S | S |
| 29 | Manipur | 243 | H | H | H | H | H | H | H | H | H | H | H | H | I | H | I | H | H | I | I | H | S | I | S |
| 30 | Delhi | 18453 | S | S | S | S | S | S | S | S | S | D | S | I | I | D | D | S | H | D | D | H | D | D | D | D | S |
| 31 | Odisha | 998 | S | S | S | H | S | H | S | S | S | S | I | I | D | S | S | S | D | I | D | D | S | D | D | D |

**Table 15: Status of States and Union Territories on July 11, 2020**

| Healthy(1) | Andaman | | | | |
|---|---|---|---|---|---|
| Infected(5) | Andhra Pradesh | Meghalaya | Puducherry | Karnataka | Arunachal |
| Symptomatic | | | | | |
| | Uttar Pradesh | Punjab | Chandigarh | Maharashtra | Telengana |
| | Jharkhand | Ladakh | Gujarat | Tamilnadu | Uttarakhand |
| | West Bengal | Himachal Pradesh | Goa | Rajastan | Jammu & Kashmir |
| | Assam | Madhya Pradesh | Chattisghar | Tripura | Kerala |
| Detected(25) | Bihar | Haryana | Manipur | Delhi | Odisha |

**Table 16: Status of States and Union Territories as on July 12, 2020**

| Healthy(4) | Infected(5) | Symptomatic(8) | Detected(14) |
|---|---|---|---|
| Telangana | Punjab | Uttarakhand | Uttar Pradesh |
| Himachala | Jharkhand | West Bengal | Andhra Pradesh |
| Meghalaya | Assam | Rajastan | Chandigarh |
| Delhi | Tripura | J&K | Maharashtra |
| | Andaman | MP | Ladakh |
| | | Karnataka | Gujarat |
| | | Chattisghar | Tamilnadu |
| | | Kerala | Goa |
| | | | Puducherry |
| | | | Bihar |
| | | | Arunachal |
| | | | Haryana |
| | | | Manipur |
| | | | Odisha |

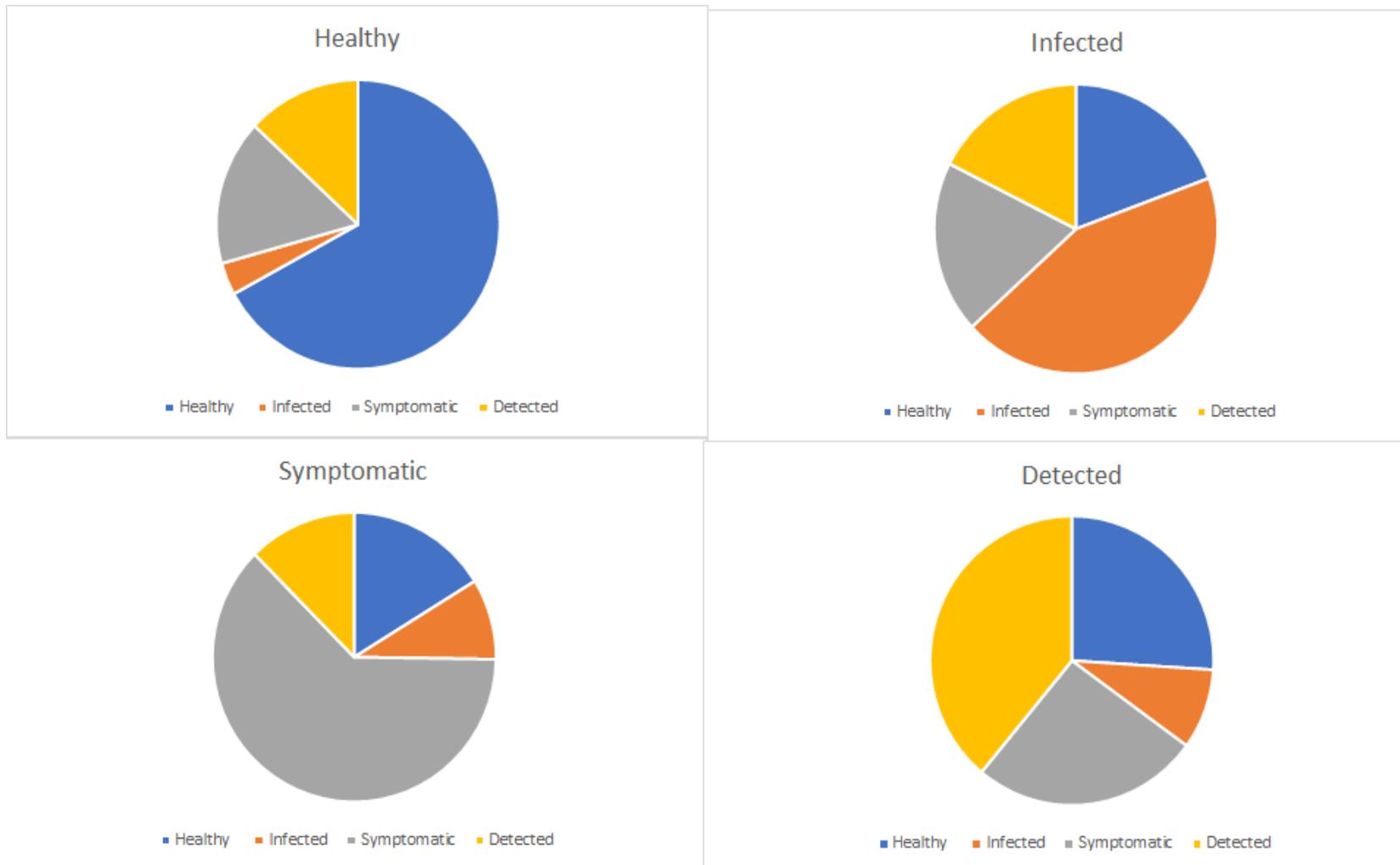

**Figure 3: State Transitions between April 30-May 20, 2020 (computed using HMM in Table 10)**

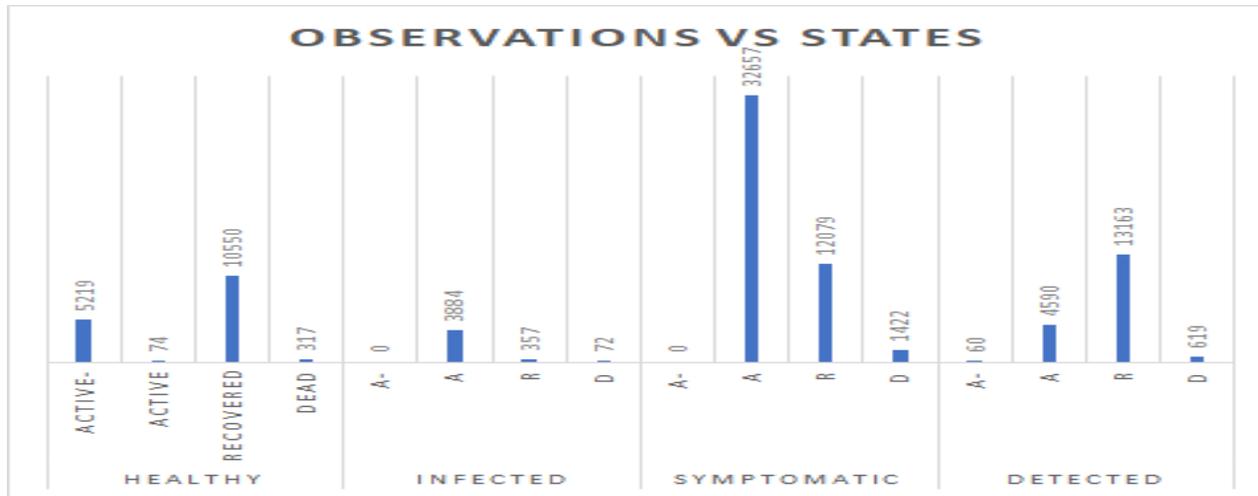

**Figure 4: Computation of Refined HMM parameter B)**

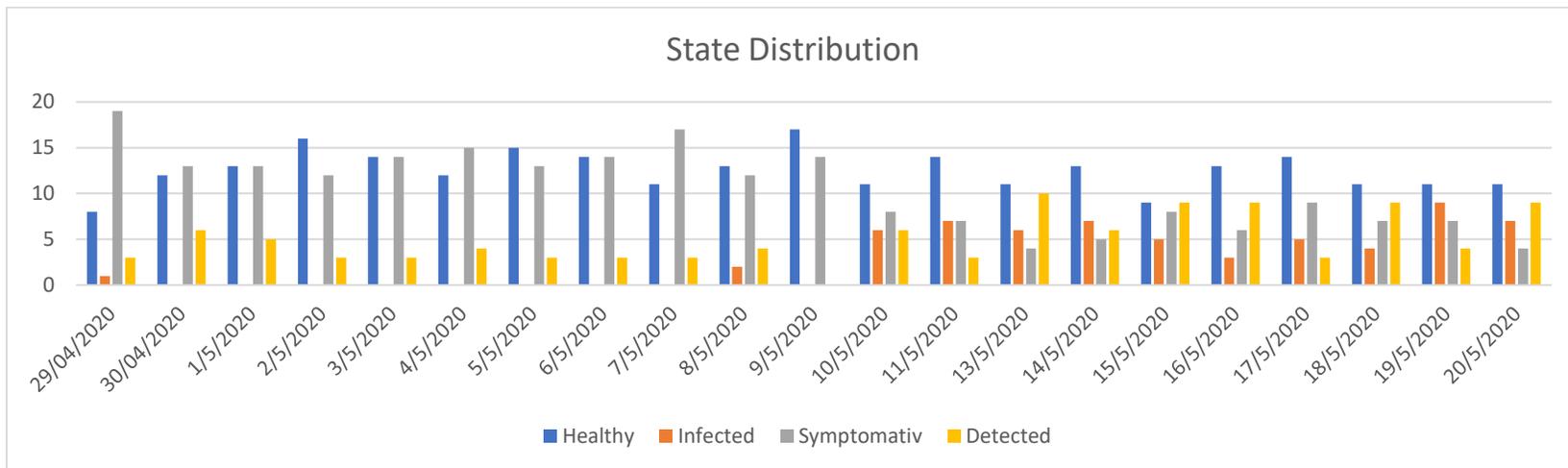

**Figure 5: Distribution of States between 29/4/2020 to 20/5/2020**

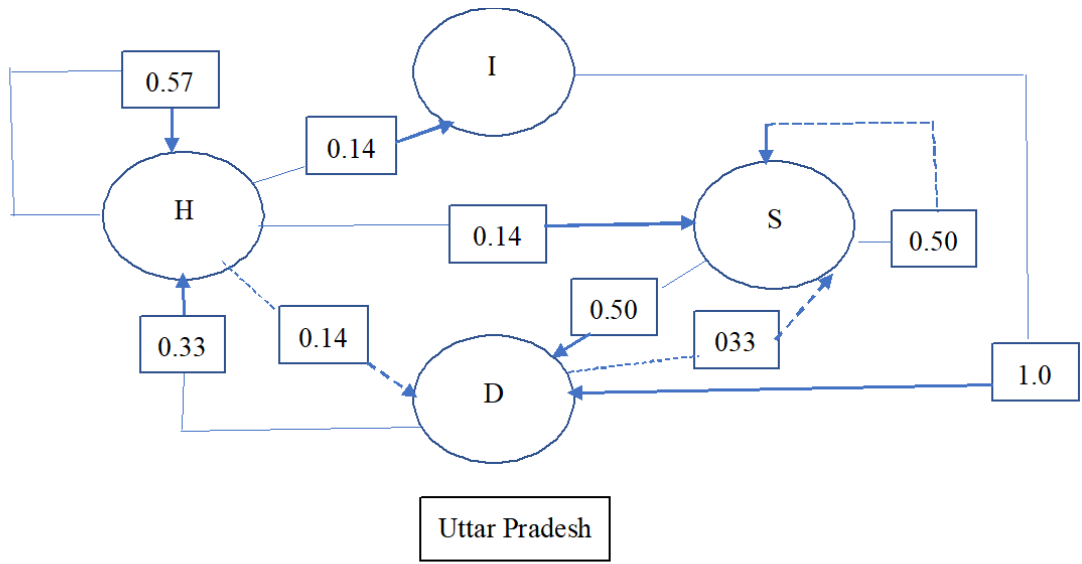
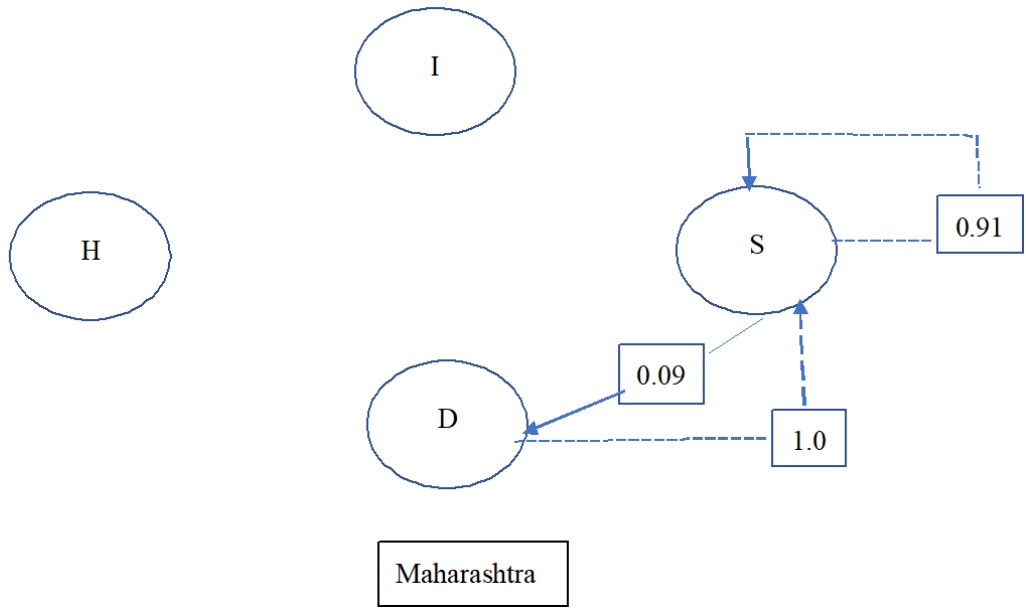

**Figure 6: Hidden State Transition Probabilities in Uttar Pradesh and Maharashtra**

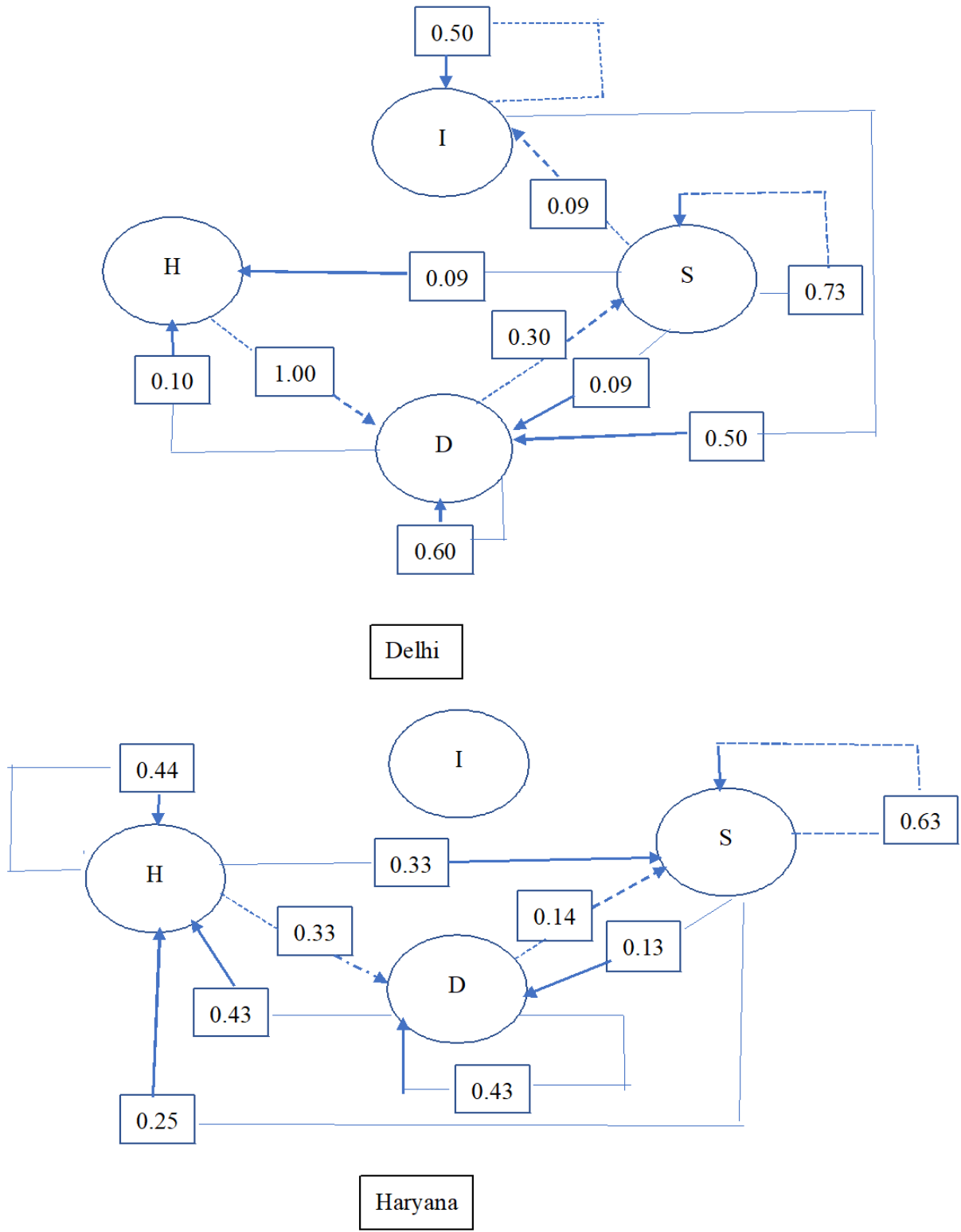

**Figure 7: Hidden State Transition Probabilities in Delhi and Haryana**

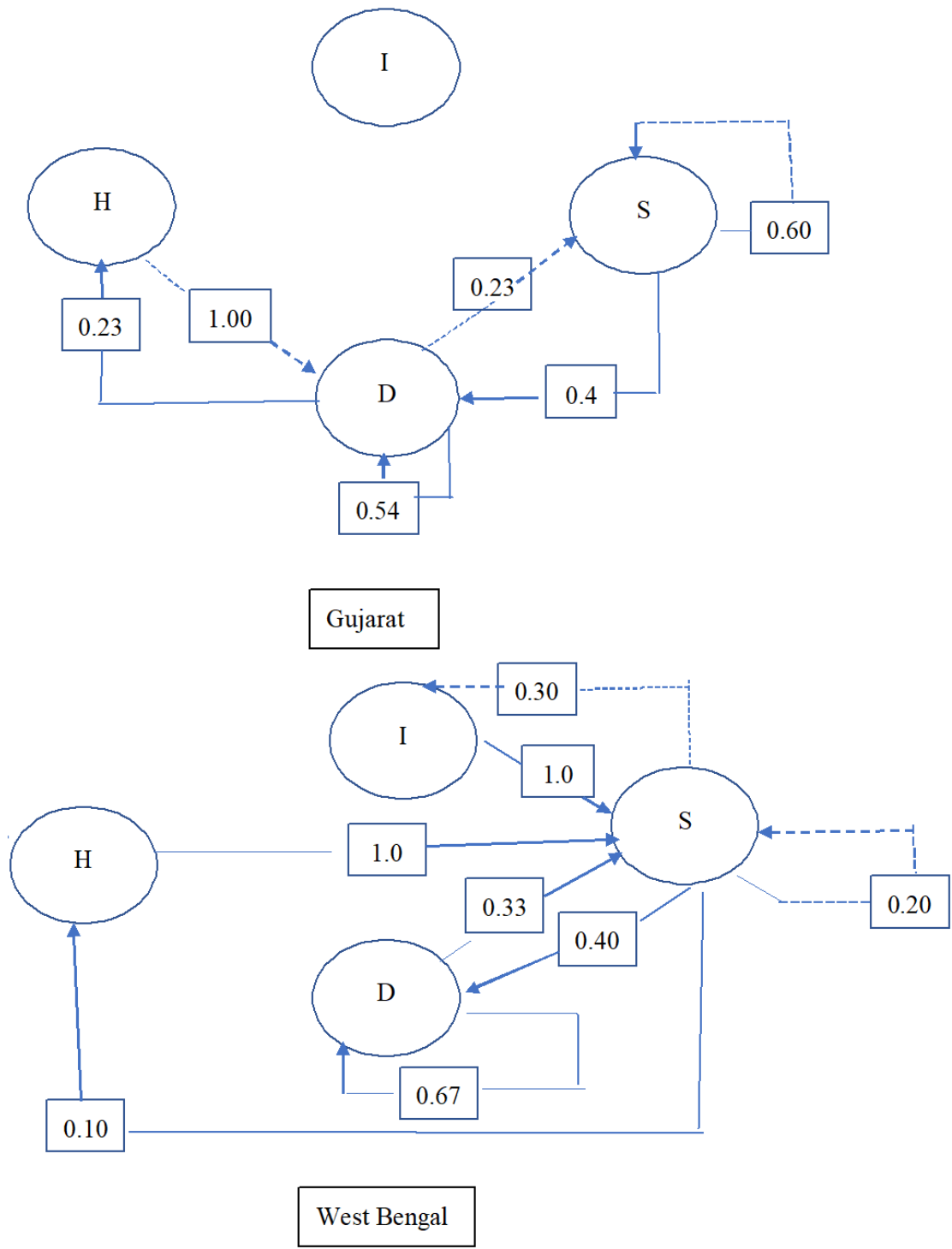

**Figure 8: Hidden State Transition Probabilities in Gujarat and West Bengal**

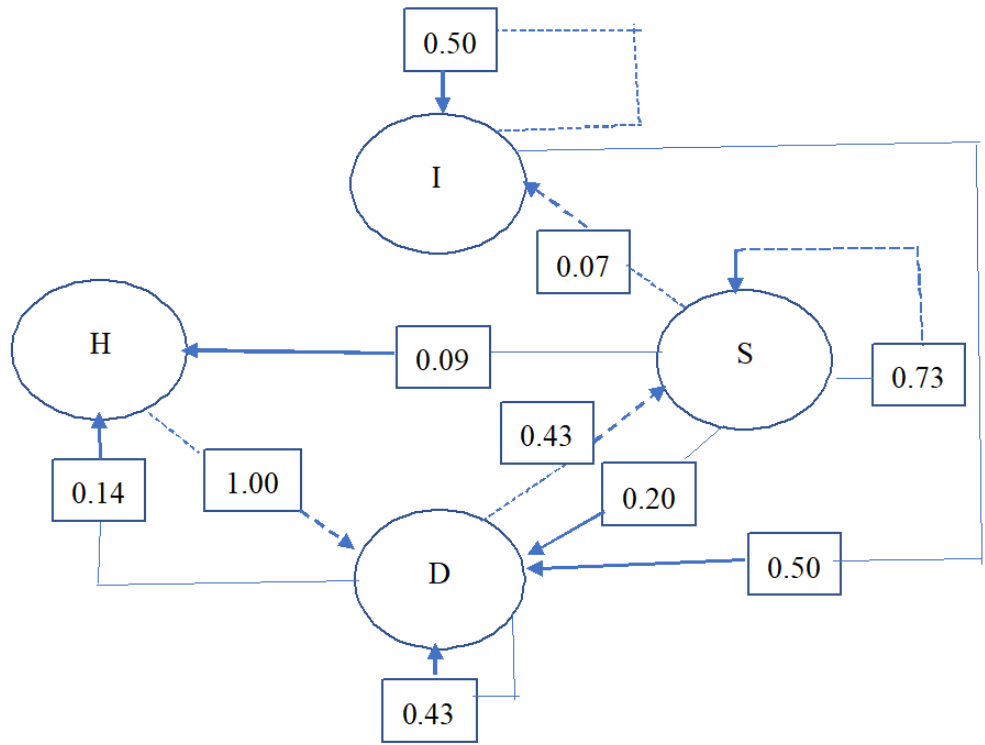

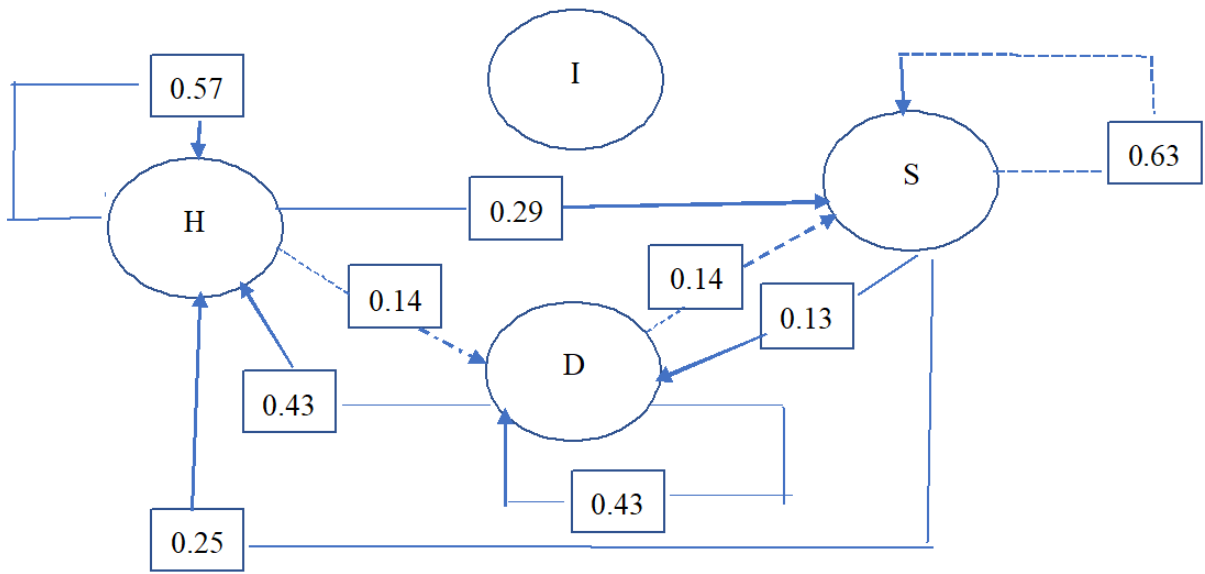

**Figure 9: Hidden State Transition Probabilities in Tamil Nadu and Karnataka**

# 6. Conclusions

The main issue unique to COVID-19 spread is that the infected may remain asymptomatic between 2 to 14 days i.e. through the incubation period. During that period, they may be unknowingly spreading the disease. Some detections may also get missed if tests return false negatives. We have proposed an approach that takes as input data reported from Hospitals in regard to active cases, recoveries and deaths and infers the latent state of regions as far as spread of COVID-19 is concerned. To that end we categorized regions as healthy, infected, symptomatic and detected (before onset of symptoms). We modeled the state of regions as hidden states and reportage from Hospitals as observations. We proposed a simpler approach to tackle the HMM learning problem in lieu of Baum Welch EM algorithm. We have also provided a set of recommendations to better manage the surveillance of COVID-19. We believe our approach can help Governments have a better assessment of spread of epidemic in different regions and enable them to plan appropriate interventions/responses.

## Summary Points

- This paper has made use of Hidden Markov Model (HMM) to assess the state of spread COVID-19 Pandemic in regions based on reportage from hospitals.
- The HMM states are categorized as Healthy, Infected, Symptomatic and Detected.
- Policy recommendations suggested for each of the above states can be calibrated further by Governments. The model can be continually refined to handle prospective and future data. Domain Experts can contribute towards validating/refining the underlying assumptions in this research.